\documentclass[conference]{IEEEtran}
\normalsize
\usepackage{cite}
\let\citep\cite 
\let\citet\cite
\usepackage{amsmath,amssymb,amsfonts}
\usepackage{algorithmic}
\usepackage{url}
\usepackage{graphicx}
\usepackage{subfig}
\usepackage{textcomp}
\usepackage{xcolor}
\usepackage{braket}
\def\BibTeX{{\rm B\kern-.05em{\sc i\kern-.025em b}\kern-.08em
    T\kern-.1667em\lower.7ex\hbox{E}\kern-.125emX}}

\usepackage{graphicx} 
\usepackage{xcolor}
\usepackage{float}
\usepackage{booktabs}
\usepackage{multirow}

\usepackage{amsmath,amsthm,stmaryrd,amsfonts,amssymb}
\usepackage{braket}

\theoremstyle{definition}

\begin{document}



\title{Mu-DisCoCat: A Variational  Pipeline for \hbox{Compositional  Generalization on Quantum Processors}\\
\begin{center}
\footnotesize
\parbox{0.95\textwidth}{
\centering
 © 2026 IEEE. Personal use of this material is permitted. Permission from IEEE must be obtained for all other uses, in any current or future media, including reprinting/republishing this material for advertising or promotional purposes, creating new collective works, for resale or redistribution to servers or lists, or reuse of any copyrighted component of this work in other works. \\ Accepted for publication at the IEEE 2nd International Conference on Quantum Artificial Intelligence (QAI 2026).}
\end{center}
}

\author{
\IEEEauthorblockN{Mina Abbaszadeh}
\IEEEauthorblockA{
Department of Computer Science \\
University College London \\
London, United Kingdom \\
m.abbaszadeh@ucl.ac.uk
}
\and
\IEEEauthorblockN{Matilda Karabina Moore}
\IEEEauthorblockA{
Department of Computer Science \\
University College London \\
London, United Kingdom \\
matilda.moore.21@ucl.ac.uk
}
\and
\IEEEauthorblockN{Raem Haq}
\IEEEauthorblockA{
Department of Computer Science \\
University College London \\
London, United Kingdom \\
raem.haq.23@ucl.ac.uk
}
\and
\IEEEauthorblockN{Martha Lewis}
\IEEEauthorblockA{
Institute for Logic, Language and Computation \\
University of Amsterdam \\
Amsterdam, Netherlands \\
m.a.f.lewis@uva.nl
}
\and
\IEEEauthorblockN{Mehrnoosh Sadrzadeh}
\IEEEauthorblockA{
Department of Computer Science \\
University College London \\
London, United Kingdom \\
m.sadrzadeh@ucl.ac.uk
}
}

\maketitle

\begin{abstract}
Achieving compositional concept generalization (CoCoGen), the ability to understand novel situations by recombining learned primitives, remains a fundamental challenge in artificial intelligence. Compositional semantic models  such as Compositional Distributional Semantics (DisCoCat) offer solutions by generalising vectors to tensors, but  suffer from  scaling bottlenecks when learning the tensors. Mapping DisCoCat onto Variational Quantum Circuits (VQCs) resolves this limitation for text, yet the methodology has not been expanded to multimodal situations such as the ones involved in CoCoGen. This paper introduces Mu-DisCoCat: a  multimodal variational quantum learning framework for DisCoCat that achieves CoCoGen. The framework first learns stable object representations from single-object image--text pairs, then fixes these and uses them to learn the relations between them in multi-object situations. In classical simulations, the model used Uhlmann state fidelity to compute the overlap between the multimodal circuit representations and achieved higher relational OOD accuracy than the evaluated CLIP baseline.  
Its deployment was evaluated using the destructive SWAP test across noisy quantum emulators, including a range of IBM fake backends, IQM FakeAphrodite, and the IBM Marrakesh quantum processor.
Despite real-world device noise, the hardware-executed models maintained a strong positive correlation with simulated fidelities, reliably distinguishing unseen similar and dissimilar pairs. Our work establishes a framework for executing CoCoGen on VQCs, demonstrating a viable use case for near-term quantum hardware.
\end{abstract}

\section{Introduction}
Humans can understand novel situations by recombining previously learned concepts, a capability known as \emph{compositional concept generalization} or CoCoGen. For example, a person who has learned that cars crossing roads are dangerous will still recognize the danger posed by a yellow car, even if they have never encountered a yellow car before. Achieving this capability in AI remains an open challenge, particularly for learning relational concepts~\cite{fodor1988connectionism,Partee1995,Smolensky1990,Pavlick2022Semantic}. Mainstream vision-language models such as OpenAI's CLIP~\cite{radford_learning_2021} achieve strong performance in image-text alignment but struggle with CoCoGen~\cite{Lewis2023,pearson2024}.  The AI models based on compositional representations of meaning should exhibit a stronger performance. One such framework is the Distributional Compositional Categorical model (DisCoCat)~\cite{Coecke2010}. Although theoretically well suited to compositional reasoning, DisCoCat has shown limited empirical success in CoCoGen,  since it represents relations as tensors,  learning which  on classical computers has a high computational cost~\cite{lewis2022clipbinding}.

DisCoCat was inspired by Categorical Quantum Mechanics \cite{abramskybob2004} and both frameworks are based on the theory of Compact Closed Categories. This enables us to represent its constructions as Variational Quantum Circuits (VQCs), allowing efficient training using hybrid quantum-classical optimization. VQCs only require a handful of parameters to learn the tensors, leading to a substantial improvement in the scalability and practicality of DisCoCat \cite{Lorenz2021,Meichanetzidis_2023}. Despite these advances, the original DisCoCat cannot be applied  to CoCoGen, as its extensions to multimodal learning are limited, e.g. \cite{nazir2024adjective} considers audio-text representations only for adjective--noun phrases and  \cite{discoclip,quclip} only focuses on concepts seen during training. 
Initial VQC-based work on CoCoGen jointly learned object and relation representations within a single-stage relational training procedure, leading to suboptimal results~\cite{ieee2025}. In our subsequent work~\cite{QCE2026}, we introduced a multistage training strategy in which object representations are first learned from single-object image-text pairs and subsequently frozen while only the relation parameters are optimised. However, that implementation relied on post-selection for the DisCoCat grammatical contractions and used larger quantum representations, including 9-qubit angle-encoded CLIP image circuits, with evaluation restricted to classical quantum simulation. In the present work, we adapt this multistage framework for near-term quantum hardware by reducing the image representation to four qubits, replacing post-selection with discard, and estimating the resulting image-caption overlap using the destructive SWAP test. This enables evaluation on noisy quantum emulators and real quantum hardware.
While Mu-DisCoCat provides an effective  model for CoCoGen, executing it on current quantum hardware presents an additional challenge. During training, the similarities between image and caption representations are computed directly from quantum states using the Uhlmann fidelity.  However, current quantum processors cannot directly access quantum states and instead provide only measurement outcomes. Consequently, deploying trained models requires a hardware compatible procedure that faithfully estimates the same similarity measure under realistic device noise. In this work, we employ the destructive SWAP test as a hardware compatible estimator of the quantum-state overlap, enabling direct experimental evaluation of the similarity learned during training. We also adapt the designed VQCs to a reduced hardware representation compatible with near-term quantum devices by reducing the depth and the number of qubits both for object and relation learning. The framework was validated on IBM Aer, multiple noisy IBM fake backends, the IBM Marrakesh quantum processor, and the IQM FakeAphrodite emulator. The results show that although hardware noise reduces the agreement with noiseless simulations, the learned similarity relationships remain positively correlated with the simulated fidelities and continue to distinguish similar from dissimilar image-caption pairs. These findings demonstrate the feasibility of achieving CoCoGen on current quantum hardware and provide a practical use case.

\section{Background}
DisCoCat encodes conceptual relations by mapping the pregroup  semantics of text to finite-dimensional Hilbert spaces via a homomorphic mapping. A pregroup is a partially ordered monoid where each element has a left and a right adjoint 
\[
{\cal P} = (P, \cdot, 1, \leq, (-)^l, (-)^r)
\]
Composition is enabled via the monoid multiplication. Whenever an element and its adjoint compose in the correct order, a cancellation happens, that is 1, the unit of our monoid multiplication. The following inequalities hold in the pregroup: 
\[
p \cdot p^r \leq 1 \qquad  p^l \cdot p \leq 1
\]
 The pregroup semantics of an atomic entity is a generator of the pregroup; that of a relation,  on the other hand, consists of multiplied adjoint types. For instance, given a pregroup generated over the set of generators $\{n,s\}$, the semantics of an atomic entity such as  ``cube''  is $n$, whereas that of a relation such as ``left of'' or ``right of'' is $n^r\cdot s \cdot n^l$. When a relation composes with two entities, the following cancellation happens: 
 \[
 n \cdot (n^r \cdot s \cdot n^l) \cdot n \leq s
 \]
 A pregroup-Hilbert Space mapping  sends atomic elements to finite dimensional Hilbert Spaces over the field of reals $\mathbb{R}$, i.e. 
 \[
 {\cal F}\colon {\cal P} \to \textbf{Hilb}
 \]
 Concretely, this mapping is defined by induction over the generators. For the above set of generators, the action of this map is defined as follows:
 \[
 {\cal F}(n) = V \qquad {\cal F}(s) = W
 \]
 It assigns a finite dimensional Hilbert space to each generator.  Pregroup elements with multiplication  and adjoints are mapped  to tensors of spaces with their duals as follows, the unit of multiplication is mapped to the field $\mathbb{R}$. Formally, for $a, b \in P$:
 \[
 {\cal F}(a^r) = {\cal F}(a^l) = {\cal F}(a)^*,  \ \ 
 {\cal F}(a \cdot b) = {\cal F}(a) \otimes {\cal F}(b),  \ \  {\cal F}(1) = \mathbb{R}
 \]
  As an example, the type of the above relational entities are mapped to the following Hilbert space:
 \[
 {\cal F}(n^r \cdot s \cdot n^l) = V^* \otimes W \otimes V^*
 \]
 The cancellations between pregroup types become as follows:
 \begin{equation}
\begin{aligned}
&{\cal F}(a \cdot a^r \leq 1) = {\cal F}(a) \otimes {\cal F}(a) ^* \rightarrow {\cal F}(1) = \mathbb{R}, \\
&{\cal F}(a^l \cdot a \leq 1) = {\cal F}(a)^* \otimes {\cal F}(a) \rightarrow {\cal F}(1) = \mathbb{R}.
\end{aligned}
\label{eq:fhilb_cups}
\end{equation}

\medskip
\noindent
Essentially, the above map takes the inner product of the elements of the tensor space, i.e. for  $v_i, v_j \in {\cal F}(a)$, we have:    
\[\sum_{i,j} c_{i,j} (v_i \otimes v_j)\mapsto \sum_{i,j} c_{i,j} \langle v_i \mid v_j \rangle\]

\noindent
This inner product helps us define matrix-vector and matrix-matrix multiplication, generalising to tensor contraction in higher orders. For the calculations that show this see~\cite{Coecke2010,coecke2013lambek}. Intuitively speaking, we are modeling semantic composition as tensor product and  semantic interactions between words as tensor contraction.   Due to space limitations we will not provide more details  and refer the reader to existing work \cite{Lorenz2021,ieee2025,Coecke2010}. 

The Hilbert space representations assigned to entities and relations are compiled into variational quantum circuits (VQCs) using parameterized ansatz. We use the Instantaneous Quantum Polynomial (IQP) ansatz~\cite{Havlicek2019QuantumFeatureSpaces} for the image circuits and the Sim4 ansatz for the textual circuits~\cite{Sim2019Expressibility}. For relational captions, Sim4 is applied to the individual word components, which are subsequently composed according to the corresponding DisCoCat grammatical reduction. The details of this assignment are given in Section~\ref{fidelity} with an example provided in Figure~\ref{discard}.  Other combinations were also considered but these  resulted in the best performance, see Table~\ref{tab:ansatz}.

\begin{table}
   \caption{Stage~1 performance of different Sim4 and IQP ansatz assignments for the image and text circuits. Validation and test splits are out of distribution (OOD), and all results are averaged over five random seeds.}
    \label{tab:ansatz}
    \centering
    \small
    \renewcommand{\arraystretch}{1.3}
    \setlength{\tabcolsep}{3pt}
    \begin{tabular}{l l c c}
        \toprule
        \textbf{Structure} & \textbf{Train} & \textbf{Valid} & \textbf{Test}\\
        \midrule
        \textbf{Image\_sim4 \& Text\_sim4} & 24.76\% & 25\% & 33.33\% \\
        \midrule
        \textbf{Image\_sim4 \& Text\_IQP} & 27.18\% & 25\% & 33.33\% \\
        \midrule
        \textbf{Image\_IQP \& Text\_IQP} & 80.54\% & 45\% & 53.34\%  \\
        \midrule
        \textbf{Image\_IQP \& Text\_sim4} & 78.66\% & 75.00\% & 66.70\% \\
        \bottomrule
    \end{tabular}
\end{table}


\section{Methods}
\subsection{Multistage Training}
%

\begin{figure}[!t]
   \centering
   \includegraphics[width=\linewidth]{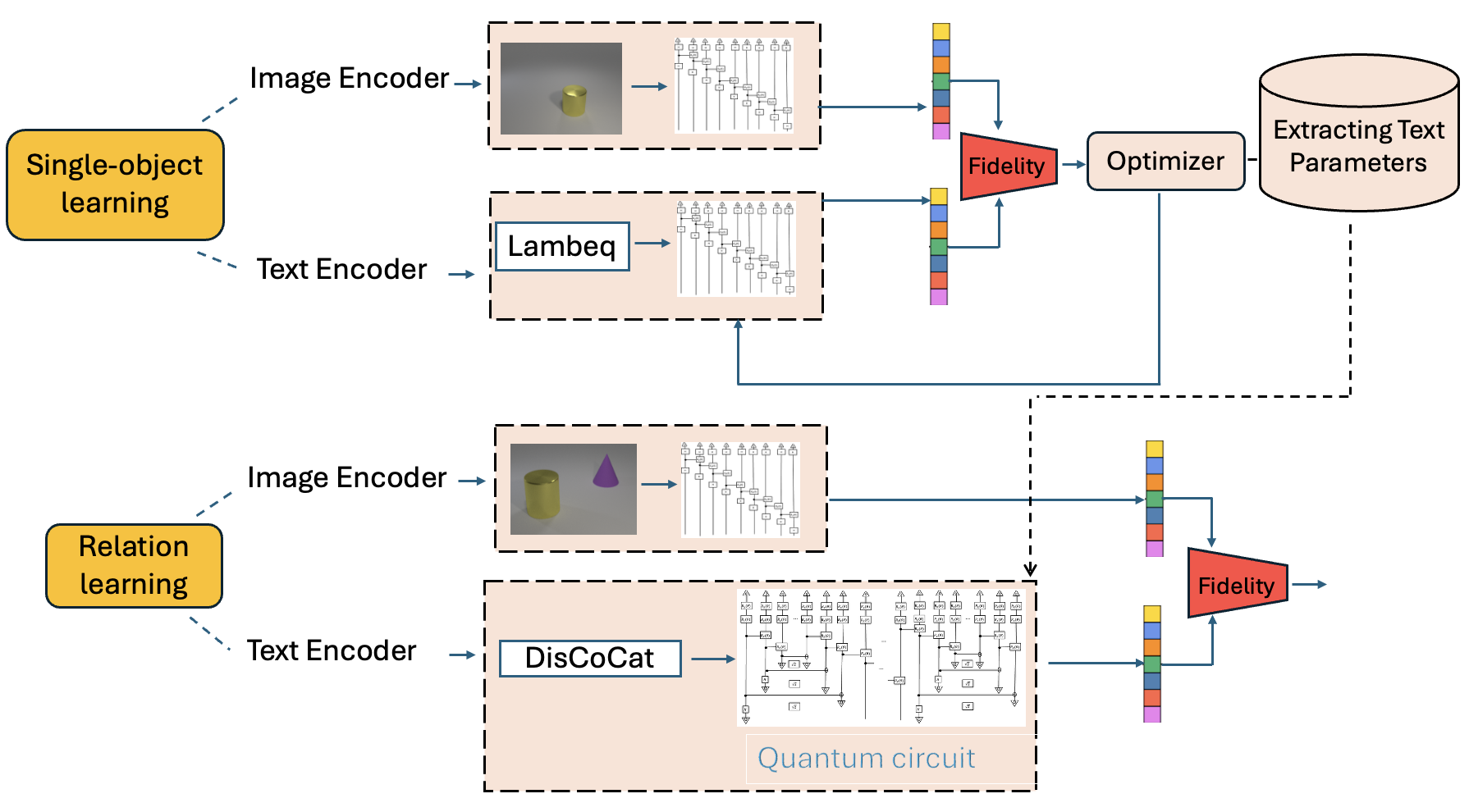}
   \caption{A multistage procedure for compositional image–caption alignment. After optimizing the noun parameters during single-object training, the learned noun parameters are transferred to relational training.}
   \label{fig:pipeline}
\end{figure}

Mu-DisCoCat consists of a two-stage training pipeline. In the first stage, object representations are learned from single-object image-caption pairs. In the second stage, the learned object parameters are frozen and transferred to relational learning, where only the relation parameters are optimized. Figure~\ref{fig:pipeline} summarizes this pipeline.

Images are turned into VQCs using several encoding strategies. For a proof-of-concept validation, we employ one-hot (OHE) and multi-hot (MHE) encodings, using five qubits per object. 
For the single-object stage, we used OHE to encode an image given the object.
While MHE is used, in the relational learning, to encode each of the two objects present in the image and their spatial relation. In both cases, each distinct type of image is represented by a binary vector and mapped onto a computational basis state, for more details see \cite{QCE2026}. 

To test for scalability on real world data, we use frozen CLIP image embeddings. These are encoded into VQCs using two standard techniques: amplitude and angle encoding. For amplitude encoding, the entire 512-dimensional CLIP embeddings are  encoded in 9 qubits using the formula
$\ket{\psi}=\sum_{i=1}^{512}x_i\ket{i}$. To keep the depth of the angle-encoded circuits low, we use a PCA-reduced version of the CLIP embeddings such that they can be encoded into the rotation angles of a quantum circuit using 9 qubits.   We also use a \textbf{Collage} encoding, where relational image circuits are constructed  manually by combining their single-object circuits using entanglement. In order to compare with a classical baseline, we use a CLIP text-projection (CTP) baseline, replacing the quantum text encoder with a CLIP text encoder while keeping the CLIP image encoder frozen. Thus, only the text-side CLIP parameters are updated during training, and the parameter counts reported for the CLIP baseline in Tables~\ref{tab:sinle-object} and~\ref{tab:relational} refer to trainable parameters rather than the total number of parameters in the full CLIP model.

In {\bf \textbf{Stage~1}}, the model is trained on single-object image--caption pairs using a supervised contrastive loss to learn grounded object representations shared across modalities.  In  {\bf \textbf{Stage~2}}, the parameters learned for each shape in the previous stage are used as fixed angles for the rotation gates on the quantum circuits and only the relational parameters (\emph{left}, \emph{right}) are optimized using an image-caption alignment.
Contrastive loss is implemented by taking  positive (similar) samples to correspond to matching image-caption
pairs, and negative (dissimilar) samples by pairing an image with an incorrect caption.


\begin{table*}[!t]
\caption{Results of single-object training}
    \label{tab:sinle-object}
    \centering
    \renewcommand{\arraystretch}{1.3}
    \setlength{\tabcolsep}{7pt}
    \begin{tabular}{l l c c c c c}
        \toprule
        \textbf{Models} & \textit{Method} & \textbf{Train (avg)} & \textbf{OOD Valid (avg)} & \textbf{OOD Test (avg)} & \textbf{Best OOD test}& \textbf{Parameters}\\
        \midrule
        \textbf{OHE} &  & 92.20\% & 95.00\% & 93.34\% & 100\% & 168\\
        \textbf{CLIP} & Angle Enc. & 78.66\% & 75.00\% & 66.70\% & 70.31\% & 312\\
                              & Amplitude Enc. &87.37\% &79.71\% & 80.97\% & 100\% & 312\\
         {\bf (Classical Baseline) CLIP} & CTP & 95.63\% & 90.45\% & 91.00\% & 95.18\% & 63.4M\\
        \bottomrule
    \end{tabular}
\end{table*}

\begin{table*}[!t]
    \caption{Results of relational learning}
    \label{tab:relational}
    \centering
    \renewcommand{\arraystretch}{1.3}
    \setlength{\tabcolsep}{7pt}
    \begin{tabular}{l l c c c c c}
        \toprule
        \textbf{Models} & \textit{Method} & \textbf{Train (avg)} & \textbf{OOD Valid (avg)} & \textbf{OOD Test (avg)} & \textbf{Best OOD Test}& \textbf{Parameters} \\
        \midrule
        {\bf MHE} & Binary & 99.00\% & 73.00\% & 85.88\% & 91.66\% & 279\\
        \textbf{Collage Encoding} 
         & Angle Enc. & 82.20\% & 61.05\% & 72.32\% & 75.00\% & 426\\
          & Amplitude Enc.& 62.28\% & 40.00\% & 48.58\% & 50.00\% & 426\\
        \textbf{CLIP Multistage} & Angle Enc. & 76.60\% & 61.25\% & 55.33\% & 60.16\% & 480\\
                               & Amplitude Enc. & 59.02\% & 45.00\% & 50.00\% & 50.00\% & 480 \\
          {\bf CLIP Non Multistage} &  Angle Enc. & 55.26\% & 44.50\% & 53.09\% & 55.00\% & 792 \\
         & Amplitude Enc.   & 50.64\% & 62.00\% & 42.93\% & 48.33\% & 792 \\
          \textbf{(Classical) CLIP} & CTP & 75.94\% & 11.14\% & 50.00\% & 50.00\% & 151.5M\\
        \bottomrule
    \end{tabular}
\end{table*}

\subsection{From Simulation to Hardware}\label{fidelity}

During \textbf{Stage~1}, both the caption and image circuits produce pure quantum states and  the similarity between them is computed using the fidelity between two pure states, as follows:
\[
F(|\psi_C\rangle,|\psi_I\rangle)
=
|\langle\psi_C|\psi_I\rangle|^2.
\]

During \textbf{Stage~2},  the sentence meanings are represented by the output $s$-wires of the quantum circuit. To isolate this output subsystem, the remaining qubits may either be post-selected or discarded. Since post-selection is prohibitively expensive on current quantum hardware, we  discard the remaining qubits. Mathematically, this corresponds to taking the partial trace over the discarded subsystem, yielding the following reduced density operator:

\[
\rho_C
=
\operatorname{Tr}_{\mathrm{discarded}}
\left(
|\Psi_C\rangle\!\langle\Psi_C|
\right).
\]

\noindent
In contrast, no qubits are discarded from the image circuit. Consequently, the image output remains a pure quantum state
$ |\psi_I\rangle,$
or equivalently,
$ \rho_I = |\psi_I\rangle\langle\psi_I|.$ The similarity between the caption and image representations is quantified during simulation using the Uhlmann fidelity. Since one argument is the pure state projector the Uhlmann fidelity reduces to the following formula:

\[
F(\rho_C,\rho_I)
=
\langle\psi_I|\rho_C|\psi_I\rangle
= 
\operatorname{Tr}(\rho_C\rho_I)
\]

\begin{figure}[!t]
\centering
    \includegraphics[width=.7\linewidth]{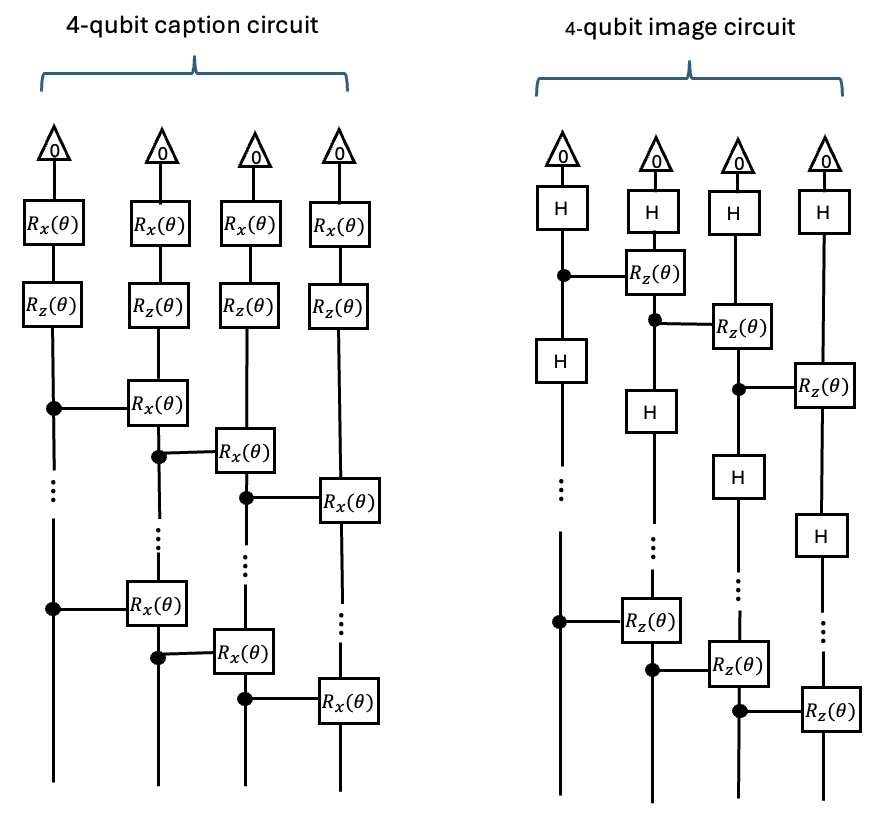}
    \caption{Four-qubit quantum circuits used in Stage 1 for single-object image–caption alignment. The caption circuit (left) is constructed using the Sim4 ansatz with trainable parameterized rotations, while the image circuit (right) uses the IQP ansatz to angle-encode PCA-reduced CLIP image features. Both circuits produce 4-qubit pure states, whose fidelity defines the image–caption similarity during training and is subsequently estimated using the destructive SWAP test.}
    \label{stage1-circuits}
\end{figure}

\begin{figure}[!t]
    \includegraphics[width=\linewidth]{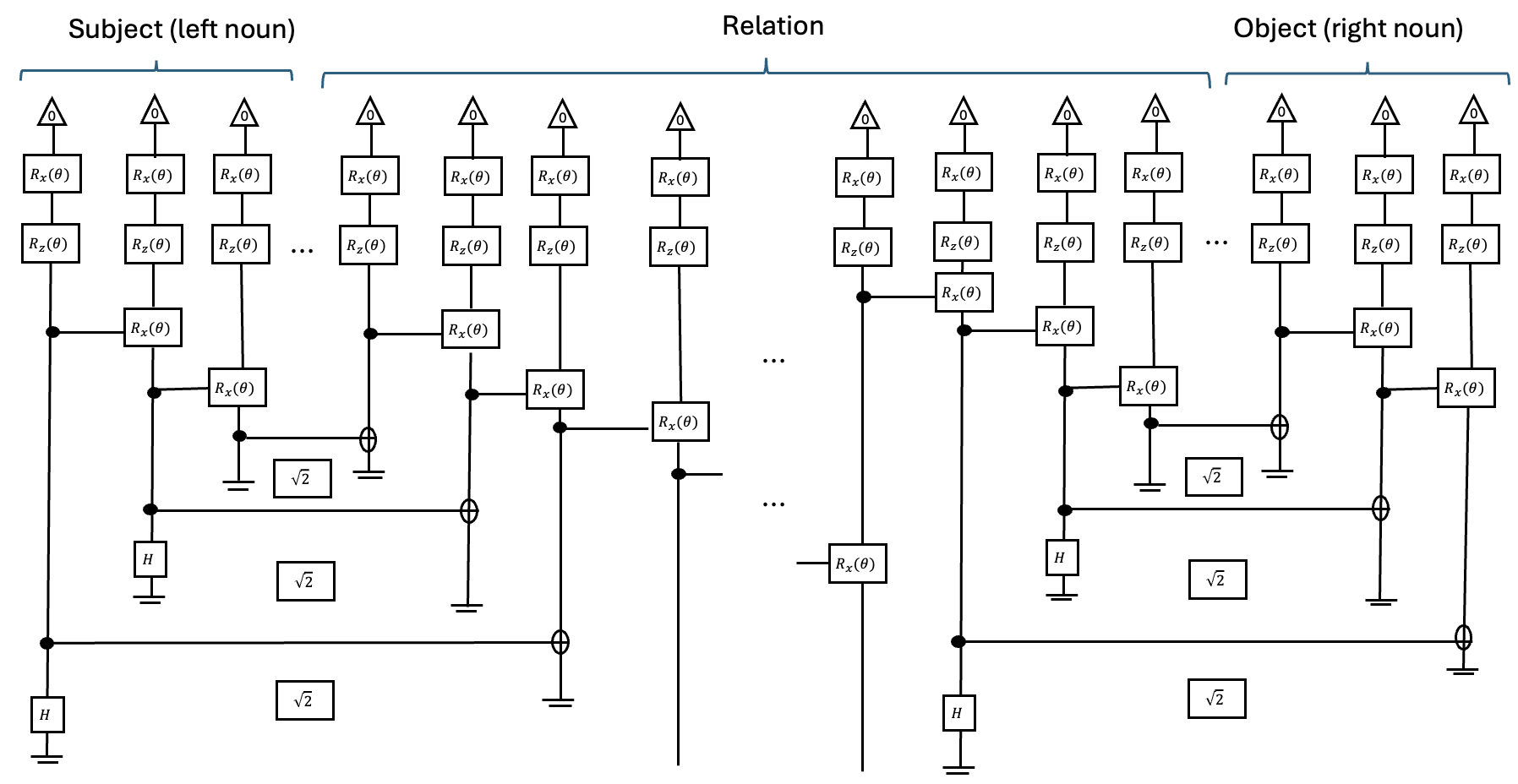}
    \caption{Variational Quantum Circuit for the relational caption obtained by applying the Sim4 ansatz to the corresponding DisCoCat diagram. The retained $s$-type output qubits represent the sentence meaning, while the remaining qubits are discarded to obtain the reduced caption density matrix $\rho_C$.}
    \label{discard}
\end{figure}

\begin{figure}[!t]
   \centering
   \includegraphics[width=\linewidth]{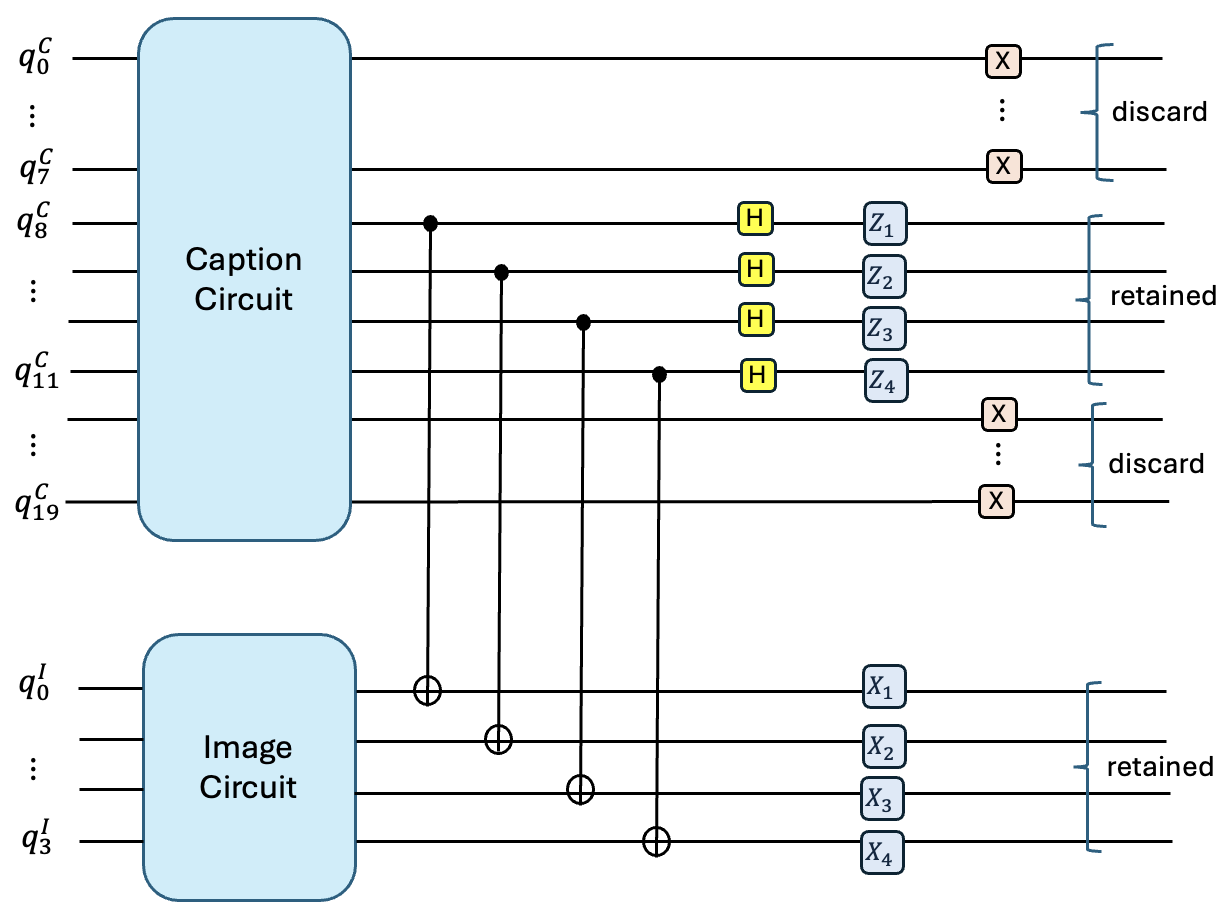}
   \caption{
    Hardware implementation of the destructive SWAP test.
    The retained output qubits of the caption circuit are compared
    with the corresponding output qubits of the image circuit using
    Bell-basis measurements.
    The measured outcomes are combined to estimate the
    Hilbert--Schmidt overlap $\mathrm{Tr}(\rho_C\rho_I)$, which
    equals the Uhlmann fidelity for a pure image state.
    The figure depicts the \textbf{Stage~2} hardware configuration;
    \textbf{Stage~1} follows the same procedure using two 4-qubit circuits.
    }
   \label{swap-test}
\end{figure}

Figure~\ref{stage1-circuits} presents the two 4-qubit circuits used in Stage~1: a Sim4 circuit for the caption representation and an IQP circuit for the image representation. Applying the same procedure to these circuits requires eight qubits in total. Figure~\ref{swap-test} illustrates the hardware implementation of the destructive SWAP test for the Stage~2 configuration. Unlike the standard SWAP test, this ancilla-free variant performs Bell-basis measurements on corresponding output-qubit pairs, thereby reducing the required qubit count. In this stage, the image branch retains a 4-qubit IQP architecture, whereas the caption branch expands to a 20-qubit circuit obtained by composing the subject noun, relation, and object noun according to the DisCoCat grammatical structure, as illustrated in Figure~\ref{discard}. The four retained output qubits of the caption circuit are paired with the image qubits, resulting in a 24-qubit hardware circuit.

\section{Simulation Evaluation}

Before validating Mu-DisCoCat on hardware, we evaluate its performance via classical simulation on an image–caption alignment task using the predefined single-object and relational splits of the CoBi2 benchmark~\cite{Pearson2025}. The benchmark consists of controlled synthetic images generated using the CLEVR rendering framework and depicts geometric shapes and spatial relations. In Stage 1, the task is to align single-object images with their corresponding object captions. In Stage 2, the task is relational image–caption alignment, where matching image–caption pairs are treated as positive samples and mismatched captions as negative samples. The benchmark is designed for out-of-distribution (OOD) compositional evaluation by assigning different combinations of the underlying concepts to the training, OOD validation, and OOD test splits. In the relational setting, the individual shapes and spatial relations are observed during training, whereas their particular combinations in the validation and test splits are unseen. The model must therefore generalize by recombining previously learned concepts.

Tables~\ref{tab:sinle-object} and~~\ref{tab:relational} summarize the performance of the original multistage models, while Table~\ref{tab:4-qubits} reports the simulation performance of the 4-qubit angle encoding model used for the subsequent hardware experiments. The motivation for introducing this reduced hardware-compatible model is discussed in Section~\ref{Hardware Validation}. All models were trained using the Adam optimizer, and the reported results are averaged over five random seeds. For the 4-qubit model, we use a learning rate of 0.009 and a batch size of 64.

\begin{table*}[!t]
\centering
\caption{Simulation Performance of the 4-qubit angle encoding model used for hardware validation.}
\label{tab:4-qubits}
\renewcommand{\arraystretch}{1.2}
\setlength{\tabcolsep}{3pt}
\begin{tabular}{l l c c c c c}
\toprule
\textbf{Models} & \textbf{Method} & \textbf{Train (avg)} & \textbf{Valid (avg)} & \textbf{OOD Test (avg)} & \textbf{Best OOD Test} & \textbf{Parameters} \\
\midrule
\textbf{Single-object training} & Angle Enc. & 94.19\% & 93.75\%  
& 93.00\% 
& 100.00\% & 132 \\
\textbf{Relational training} & Angle Enc. & 78.91\% & 77.46\% & 72.69\% & 73.84\% & 210 \\
\bottomrule
\end{tabular}
\end{table*}

Overall, our framework consistently improves relational compositional generalization while using substantially fewer trainable parameters than both the non-multistage quantum models and the classical CLIP baseline. For \textbf{single-object} training, \textbf{OHE} achieves the highest quantum accuracy (93.34\%), while amplitude encoding performs best among the CLIP-based quantum models (80.97\%),  using only 312 trainable parameters compared with 63.4M for CLIP. For \textbf{relational} learning, the classical CLIP baseline performs at chance level (50\%), whereas all multistage quantum models achieve substantially higher accuracies using only a few hundred trainable parameters versus 151.5M for CLIP, with \textbf{MHE} obtaining the best overall result (85.88\%) and the  \textbf{Collage} encoding following suit at 72.32\%. Among the CLIP-based quantum models, amplitude encoding was more effective for single objects, whereas angle encoding consistently yielded better relational generalization. It is important to note that the quantum models and the CLIP baseline differ substantially in model capacity. The comparison is therefore not parameter-matched and should not be interpreted as evidence of computational or quantum advantage. Rather, the results show that the proposed structured quantum models can achieve competitive, and in the relational OOD setting higher, compositional generalization performance with a much smaller trainable parameter space. This suggests that the compositional structure and multistage training strategy may provide a useful inductive bias for this task. However, the smaller number of trainable parameters does not necessarily imply lower computational cost on current quantum hardware, where circuit execution, sampling, and noise introduce additional overheads.

\begin{figure}[t]
\centering
\subfloat[Amplitude-Encoding\label{fig:amplitude_enc}]{
    \includegraphics[width=.4\linewidth]{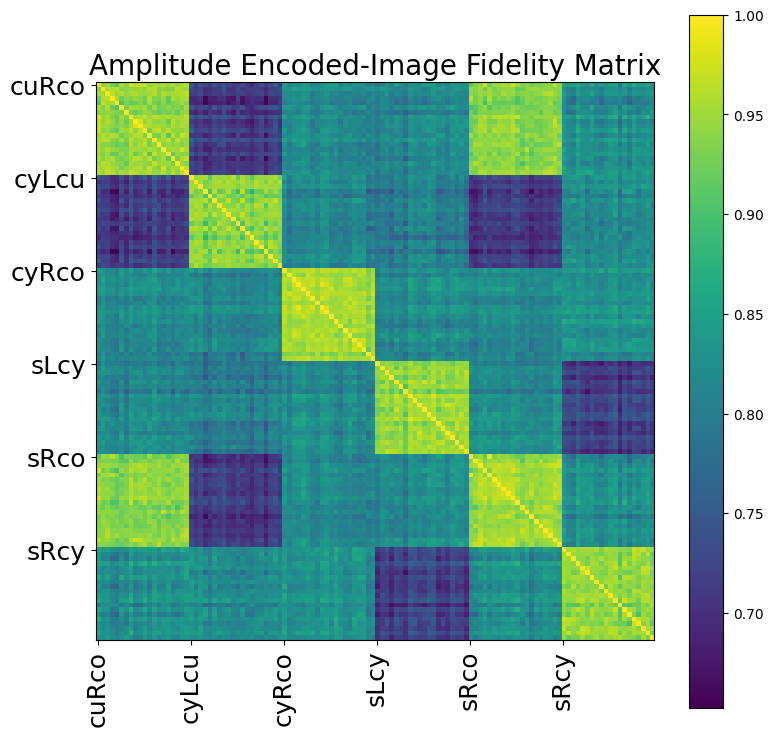}
}\hspace{0.02\linewidth}
\subfloat[Angle-Encoding\label{fig:angle_enc}]{
    \includegraphics[width=.4\linewidth]{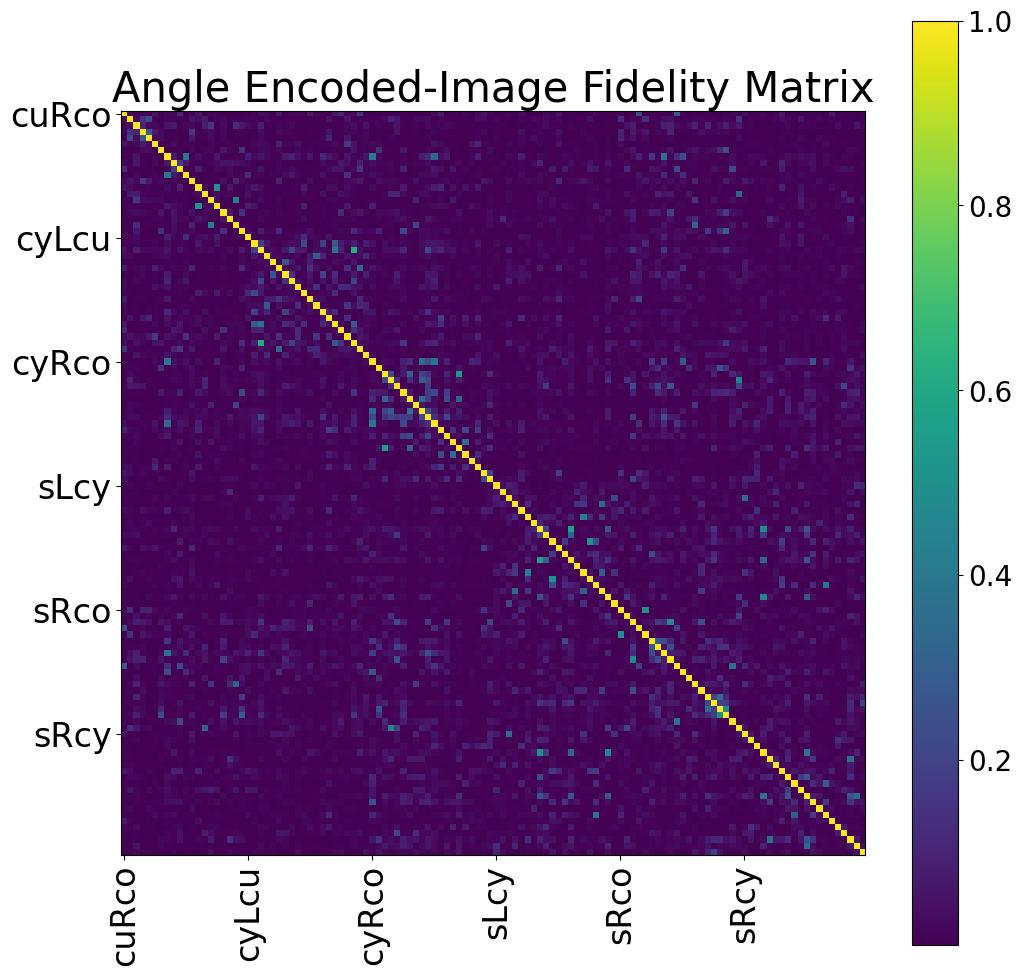}
}
\caption{Amplitude vs. angle encoding state fidelity on images from the test dataset. The desired pattern is block diagonal, indicating high similarity within class and lower similarity between classes. Captions are abbreviated as: \textit{cube right cone} (cuRco), \textit{cylinder left cube} (cyLcu), \textit{cylinder right cone} (cyRco), \textit{sphere left cylinder} (sLcy), \textit{sphere right cone} (sRco), and \textit{sphere right cylinder} (sRcy).}
\label{fig:encodings}
\end{figure}

We hypothesize that the nonlinear rotations of angle encoding produce more separable relational representations, contributing to an improved performance (see Figure~\ref{fig:encodings}).
Furthermore, multistage training consistently outperforms the corresponding  non multistage  models while reducing the number of trainable parameters (480 vs.~792), supporting the hypothesis that learning stable object representations before relational reasoning improves compositional generalization. Fine-tuning the CLIP image encoder did not improve the relational performance, consistent with previous findings~\cite{Lewis2023}.

Table~\ref{tab:4-qubits} shows that the reduced 4-qubit angle-encoding model maintains strong simulation performance. In particular, the relational model achieves 72.69\% OOD test accuracy using only 210 trainable parameters, demonstrating that the reduced hardware-compatible model provides a suitable basis for the hardware validation presented in Section~\ref{Hardware Validation}.

\section{Hardware Validation}\label{Hardware Validation}

In order to run Mu-DisCoCat on hardware, we need to make amendments. For text, we use DisCoCat circuits with discard, as described in Section III. For images, we focus on angle encoding because it uses a PCA-reduced CLIP representation and exhibits a clearer class-separation structure in the image-state fidelity analysis compared with amplitude encoding (see Figure~\ref{fig:encodings}). To keep the number of resources low, we train a  4-qubit angle-encoding model using a PCA-based image representation (see Table~\ref{tab:4-qubits}). The corresponding \textbf{Stage~1} and \textbf{Stage~2} hardware circuit configurations are as described in Section~\ref{fidelity}. Consistent with the multistage training framework, the noun parameters learned during single-object training remain fixed during the relational stage, while the relation parameters are optimized.

For hardware evaluation, state fidelity cannot be accessed directly. This is because current quantum hardware returns measurement outcomes and repeated circuit executions yield bit strings sampled from the Born rule probability distribution. These measurement statistics do not directly provide the quantum fidelity used during simulation. Instead, the two quantum states must be compared jointly, which is achieved here using the destructive SWAP test~\cite{BuhrmanCleveWigderson2001,Cincio2018}, which  estimates the Hilbert--Schmidt overlap
$\operatorname{Tr}(\rho_C\rho_I)$.
We know
\[
\operatorname{Tr}(\rho_C\rho_I)
=
\frac{1}{N}
\sum_{k=1}^{N}
\prod_{i=1}^{4}
(-1)^{z_i^{(k)}x_i^{(k)}},
\]
where $N$ is the number of shots and
$z_i^{(k)},x_i^{(k)}$ denote the Bell-measurement outcomes
of the $i$th output-qubit pair in shot $k$.
Therefore, the destructive SWAP test directly estimates the same similarity measure optimized during simulation, without requiring quantum state tomography.





The  setting was tested on hardware by executing the above described circuits, after training,  on the  held-out test set of the benchmark. To evaluate the performance, we  compared  the simulated test fidelities with the hardware estimated destructive SWAP test overlaps. This allowed  us to assess how faithfully the learned similarity relationships are preserved after deployment.

\begin{figure*}[t!]
\centering
\subfloat[IBM Aer (Noiseless)\label{fig:ibm_aer}]{
    \includegraphics[width=.32\linewidth]{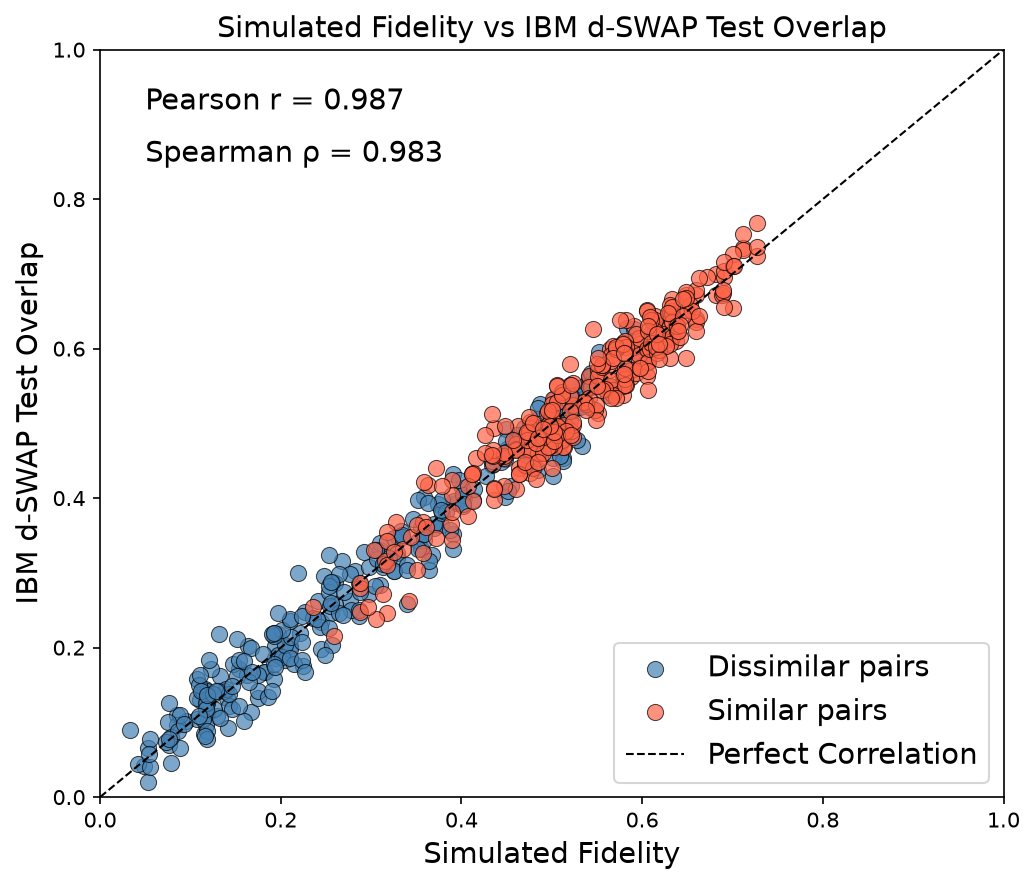}
}\hfill
\hfill
\subfloat[IBM FakeMarrakesh (Noisy)\label{fig:ibm_fmar}]{
    \includegraphics[width=.3\linewidth]{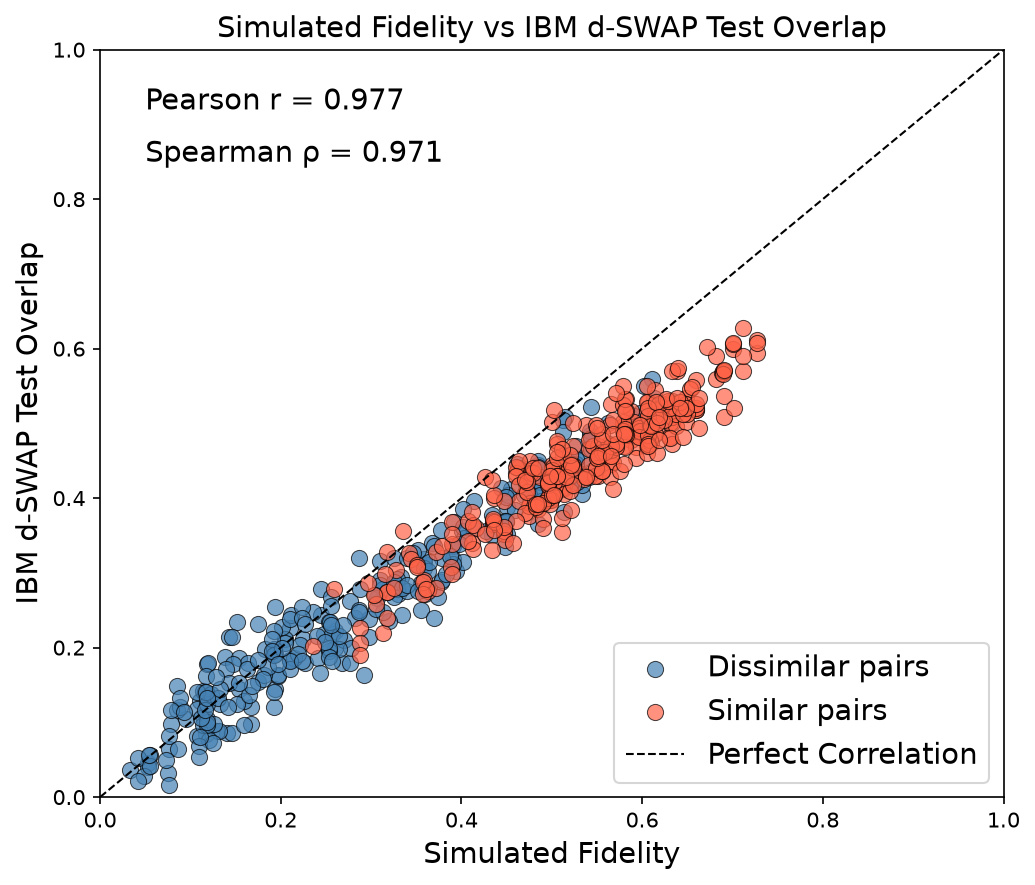}}
\hfill    
\subfloat[IBM Marrakesh (Real Hardware)\label{fig:ibm_mar}]{
    \includegraphics[width=.3\linewidth]{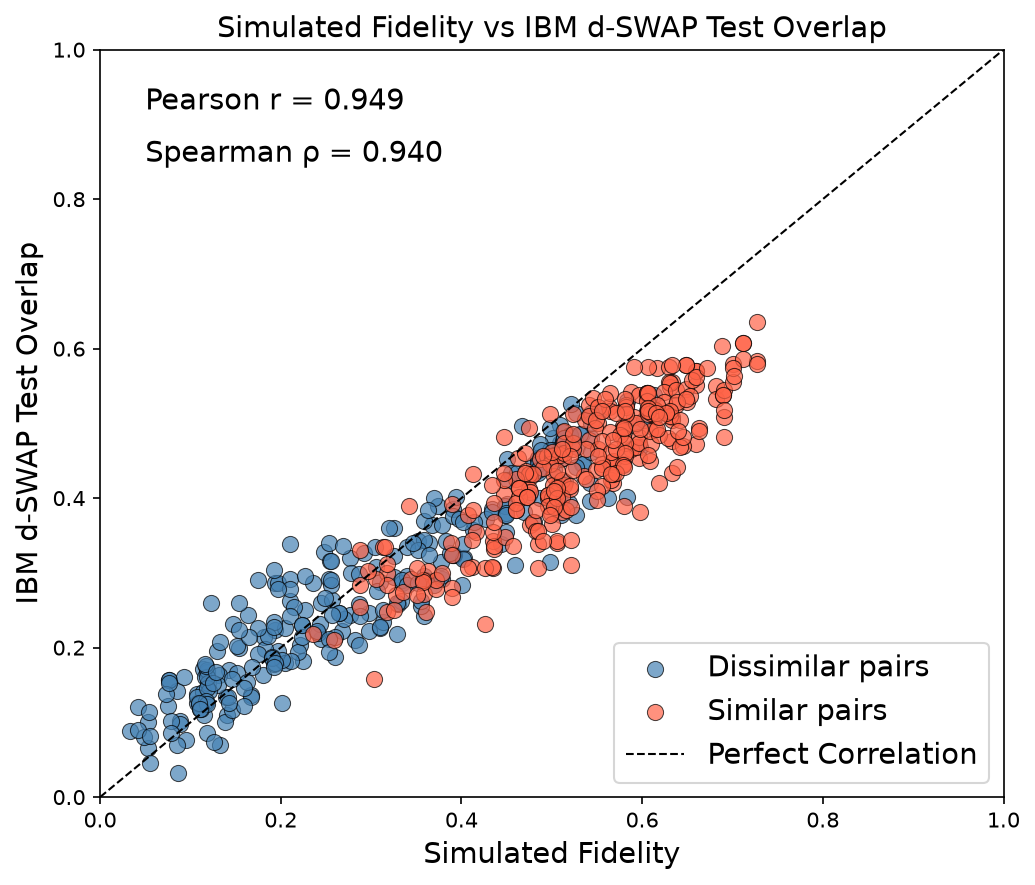}
}
\caption{\textbf{Stage~1} simulated fidelity versus destructive SWAP test overlap across quantum emulators and hardware. Simulated fidelity is evaluated on the held-out test pairs using the classically trained model. Red points denote similar (correct) image--caption pairs, while blue points denote dissimilar (incorrect) image--caption pairs.}
\label{fig:pearson-stage1}
\end{figure*}

Figure~\ref{fig:pearson-stage1} shows that for the \textbf{single-object} model, the destructive SWAP test overlaps closely follow the simulation fidelities across both quantum emulators and real hardware. Although hardware noise slightly increases the scatter around the ideal diagonal, similar and dissimilar image--caption pairs remain well separated, indicating that
the learned similarity relationships are faithfully preserved. 

\begin{figure*}[!t]
\centering
\subfloat[IBM Aer (Noiseless)\label{fig:ibm_aer}]{
    \includegraphics[width=.3\linewidth]{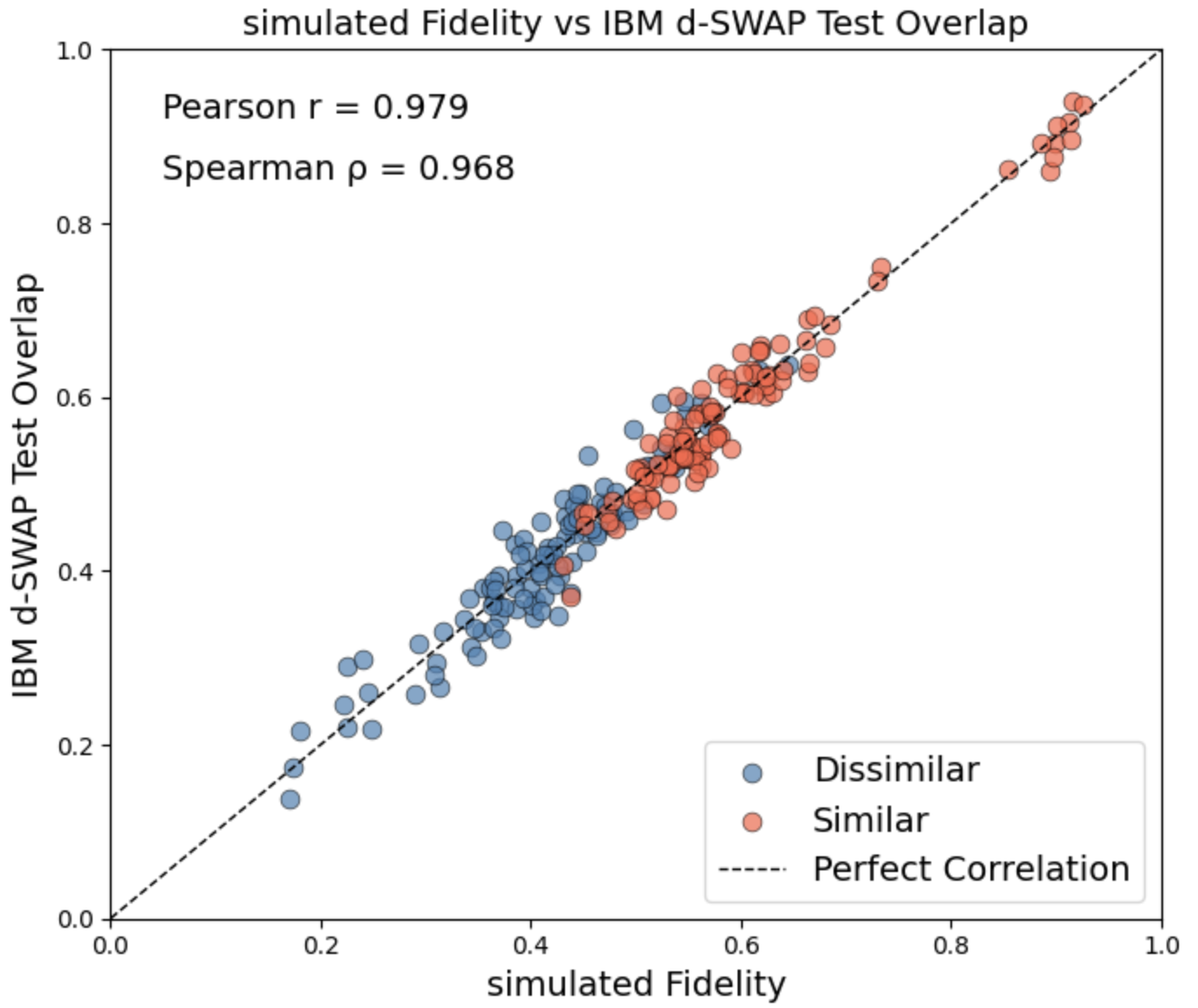}
}\hfill
\subfloat[IBM FakeMarrakesh (Noisy)\label{fig:ibm_fm}]{
    \includegraphics[width=.3\linewidth]{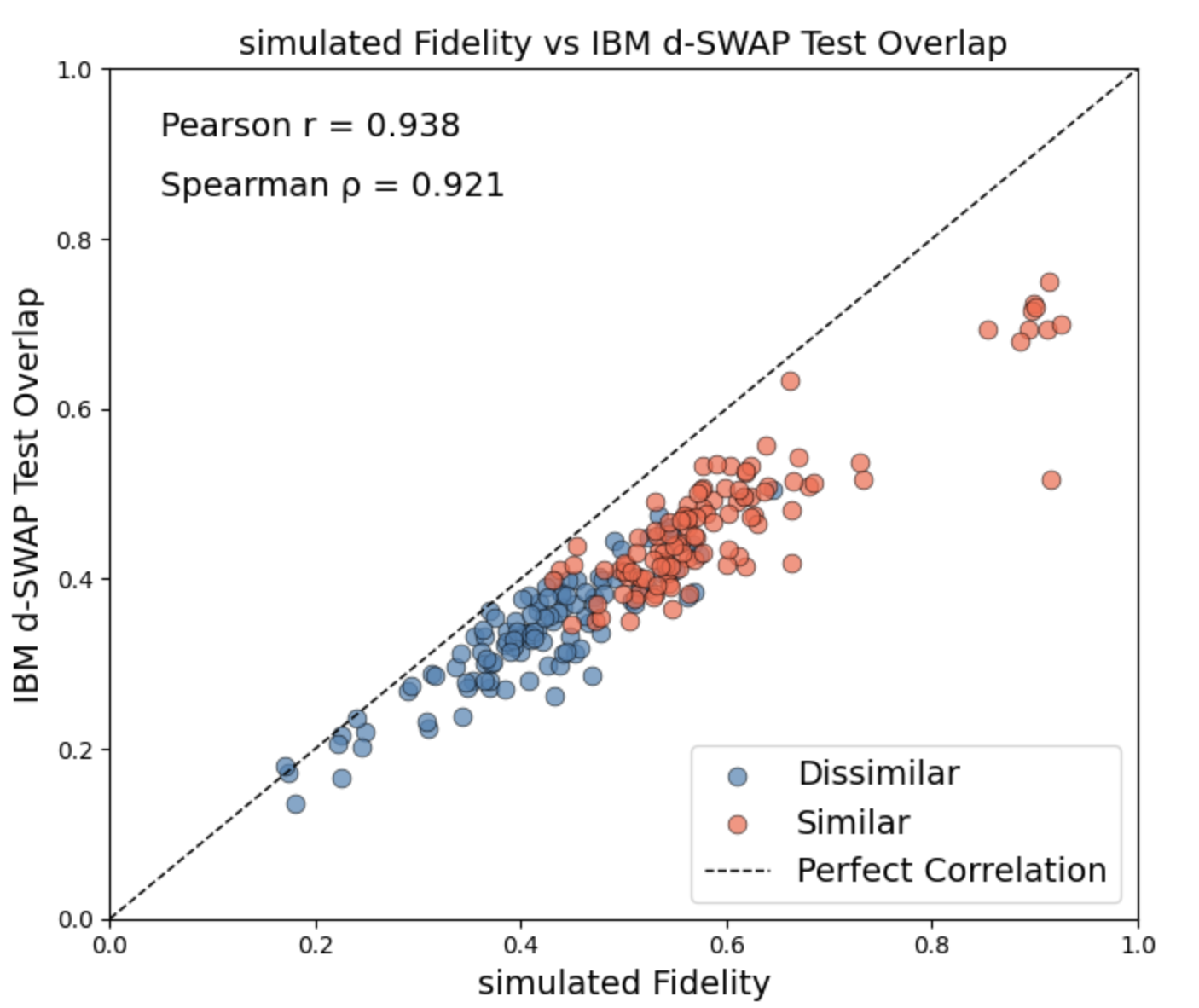}
}\hfill
\subfloat[IBM Marrakesh (Real Hardware)\label{fig:ibm_mar}]{
    \includegraphics[width=.3\linewidth]{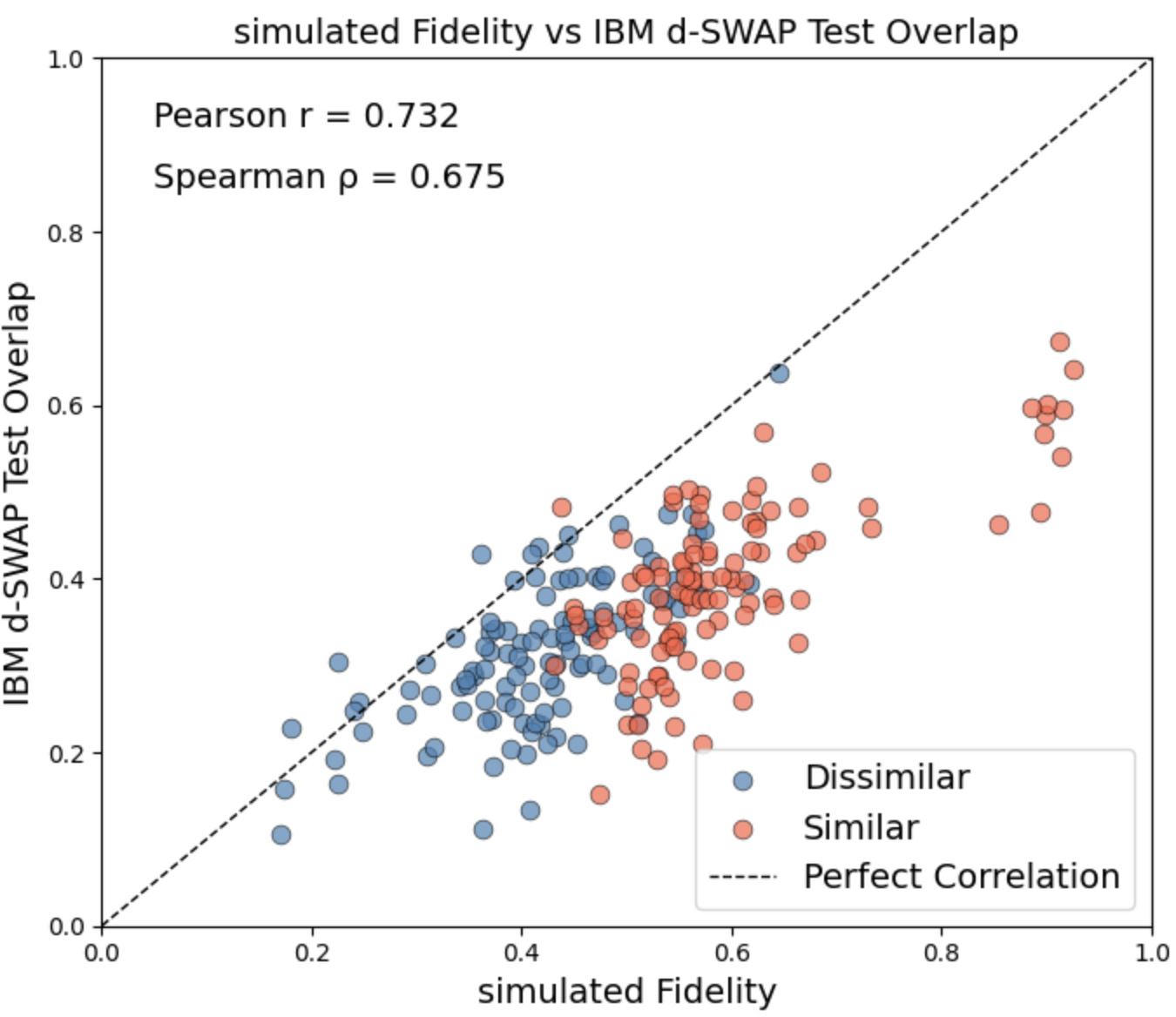}
}
\caption{\textbf{Stage~2} simulated fidelity versus destructive SWAP test overlap across quantum emulators and hardware. Simulated fidelity is evaluated on the held-out test pairs using the classically trained model. Red points denote similar (correct) image–caption pairs, while blue points denote dissimilar (incorrect) image–caption pairs.}
\label{fig:pearson-stage2}
\end{figure*} 

For \textbf{relational} learning, Figure~\ref{fig:pearson-stage2} illustrates excellent agreement between the destructive SWAP test estimates and the exact fidelities in the noiseless \texttt{IBM Aer} and  noisy \texttt{IBM FakeMarrakesh} emulators. In contrast, agreement decreases on \texttt{IBM Marrakesh}, illustrating the impact of hardware noise and device constraints.

In  \textbf{Stage~2}, slightly stronger agreement was observed for \texttt{IBM FakeMontreal} than for \texttt{IBM FakeMarrakesh}. This is consistent with the differences in the respective backend calibration. Although both transpiled circuits use 24 active qubits, the average transpiled depth is marginally higher for \texttt{IBM FakeMarrakesh} (365 versus 346), which may contribute to the accumulation of gate errors during execution.

\begin{table*}[h!]
\centering
\caption{Comparison of destructive SWAP test overlap estimates and classical quantum-state fidelities across 580 image-caption pairs for single objects. Mean overlaps, Pearson correlation, centroid distance and mean absolute error are reported.}
\label{tab:emulator_results_stage1}
\begin{tabular}{lcccccc}
\toprule
\textbf{Configuration} & \textbf{Dissimilar} & \textbf{Similar} & \textbf{$r$ with classical}&\textbf{Centroid Distance}&\textbf{MAE} \\
\midrule
 Single-Object Training Model (Classical)& $0.298\pm 0.155$ & $0.533 \pm 0.103$ & $-$ & $0.234$ &$-$\\
IBM Aer (Noiseless),  1000 shots & $0.297 \pm 0.107$ & $0.532 \pm  0.107$ & $0.987$ & $0.234$ & $0.023$\\
 IBM Aer (Noiseless), 100 shots  & $ 0.300\pm 0.179$ & $0.536\pm 0.134$ & $0.888$ & $0.235$& $0.073$\\
  IQM FakeAphrodite (Noisy), 1000 shots & $0.093 \pm 0.034$ & $0.106 \pm 0.032$ & $0.294$ & $0.012$ & $0.316$\\
  IBM FakeMarrakesh (Noisy), 1000 shots & $0.261 \pm 0.129$  & $0.446\pm 0.083$ & $0.977$ & $0.185$ & $0.067$\\
  IBM Marrakesh (Real Hardware), 1000 shots & $0.282 \pm  0.124$ & $0.443 \pm 0.089$ & $0.949$ & $0.161$ & $0.069$\\
\bottomrule
\end{tabular}
\end{table*}

\begin{table*}[h!]
\centering
\caption{Comparison of destructive SWAP test overlap estimates and classical quantum-state fidelities across 260 image--caption pairs for relational circuits. Mean overlaps, Pearson correlation, centroid distance and mean absolute error are reported.}
\label{tab:emulator_results_stage2}
\begin{tabular}{lcccccc}
\toprule
\textbf{Configuration} & \textbf{Dissimilar} & \textbf{Similar} & \textbf{$r$ with classical}&\textbf{Centroid Distance}&\textbf{MAE} \\
\midrule
Relational Training Model (Classical)& $0.413\pm0.091$& $0.594\pm0.113$ & $-$ & $0.181$ &$-$\\
IBM Aer (Noiseless), 1000 shots & $0.414 \pm 0.098$ & $0.592 \pm 0.118$ & $0.979$ & $0.178$ &$0.023$\\
 IBM Aer (Noiseless), 100 shots  & $0.432 \pm 0.129$ & $0.602 \pm 0.136$ & $0.803$ &$ 0.170$ &$0.063$\\
  IBM FakeMontreal (Noisy), 1000 shots & $0.364 \pm 0.074$ & $0.494 \pm 0.092$ & $0.968$ &$0.130$ &$0.076$\\
  IQM FakeAphrodite (Noisy), 1000 shots &$0.175 \pm 0.035$&$0.206 \pm 0.038$& $0.585$ &$0.031$ &$0.313$\\
  IBM FakeMarrakesh (Noisy), 1000 shots &  $0.336\pm 0.070$ &  $0.471\pm 0.088$  & $0.938$ & $0.135$ &$0.099$\\
  IBM Marrakesh (Real Hardware), 1000 shots & $0.312 \pm 0.086$ & $0.395 \pm 0.086$ & $0.732$ & $0.083$ &$0.159$\\
\bottomrule
\end{tabular}
\end{table*}

Consistent with the figures, see Table~\ref{tab:emulator_results_stage1} for the exact correlation degrees between   simulation fidelities in  the single-object model across both quantum emulators and real hardware. The  reported strong agreement is expected because \textbf{Stage~1} compares two 4-qubit quantum states using an 8-qubit destructive SWAP test circuit, resulting in a relatively shallow and compact hardware implementation that is less susceptible to accumulated noise.  In line with this observation, Table~\ref{tab:emulator_results_stage1} also shows relatively small mean absolute errors (MAE) and only a modest reduction in centroid distance across the noisy backends, indicating that the similarity structure learned during training remains largely intact despite hardware imperfections.

The hardware validation results are consistently stronger for single-object training than for relational training. As described in Section~\ref{fidelity}, \textbf{Stage~1} compares two 4-qubit quantum states using an 8-qubit test circuit, whereas \textbf{Stage~2} requires a 24-qubit hardware circuit due to the larger relational caption representation. This increase in circuit complexity is reflected in the transpiled circuits, whose average  depth increases from 16 in \textbf{Stage~1} to 365 in \textbf{Stage~2}. Consequently, the relational circuits are substantially more susceptible to noise accumulation during execution.  Table~\ref{tab:emulator_results_stage2} also shows that increasing the number of shots improves agreement with the classical fidelities.

The \texttt{IQM FakeAphrodite} backend produces lower overlap values and weaker correlation than the IBM emulators. Inspection of the backend calibration data suggests that the stronger degradation is consistent with shorter median coherence times ($T_1 \approx 48.9$ us and $T_2\approx 9.60$ us) together with higher gate (1-qubit gates: $\approx 0.001$, 2-qubit gates: $\approx 0.015$) and readout error rates ($\approx 0.05$) compared to the IBM backends.
Nevertheless, all backends retain a positive correlation (Pearson $r$) with the classical fidelities. Naturally, executions on the real hardware yielded the lowest correlations but as plot (c) of Figure~\ref{fig:pearson-stage2} shows, similar and dissimilar image-caption pairs remain distinguishable. Together, the cluster centroid distance and the MAE indicate that, although agreement with the simulated fidelities gradually decreases from noiseless simulation to real hardware, the separation between similar and dissimilar image-caption pairs is preserved.

\section{Conclusions and Future Work}
We presented a theoretical framework for a multimodal compositional semantic model (Mu-DisCoCat), where meanings of words are represented in Hilbert Space and encoded as VQCs. Meanings of sentences are obtained by composing these using tensor contraction, encoded as Bell measurements in VQCs. Meanings of images are also VQCs, obtained by either directly encoding the image vectors as VQCs (CLIP, One-Hot, Multi-Hot), or by concatenating the VQCs of each object within the image (Collage). We provided a hardware-compatible implementation of Mu-DisCoCat by replacing post-selection with discard and state overlap fidelity with the destructive SWAP test. The model was trained and tested on a Compositional Concept Generalisation (CoCoGen) benchmark. Classical simulations show better results over mainstream AI, in particular OpenAI's CLIP. Although hardware noise reduced the quantitative agreement with the simulated fidelities, the framework kept a strong correlation with the simulated fidelities. In particular, similar and dissimilar image-caption pairs remain clearly distinguishable across all evaluated hardware platforms, demonstrating that the proposed  framework preserves the learned similarity relationships after deployment on current quantum processors. These results show that Mu-DisCoCat can preserve compositional similarity relationships when deployed on current quantum processors. This supports its practical deployment  on current quantum hardware for future use on other similar applications. 

One shortcoming is that we only trained the text component and kept the visual representations fixed. This is  due to the lack of a compositional distributional theory for images. Future work will focus on constructing such a theory, translating it into VQCs, and evaluating compositional concept generalization on larger real-world datasets. Another important direction is to investigate quantum error mitigation techniques to improve the robustness of destructive SWAP test measurements on near-term quantum hardware, as well as evaluating the proposed framework on future fault-tolerant quantum devices.

\bibliographystyle{IEEEtran}   
\bibliography{references_consistent}

\end{document}